\documentclass[letterpaper, 10 pt, conference]{ieeeconf}  

\IEEEoverridecommandlockouts    

\usepackage{graphics} 
\usepackage{amsmath} 
\usepackage{amssymb}  
\usepackage{graphicx} 
\usepackage{multirow} 
\usepackage[labelformat=simple]{subcaption}
\usepackage{xcolor}
\usepackage{mathtools}
\usepackage[ampersand]{easylist}
\usepackage{boldline}
\usepackage{booktabs}
\usepackage{adjustbox}
\usepackage{mathtools}
\usepackage{algorithm}
\usepackage[noend]{algpseudocode}
\usepackage{hyperref}
\usepackage{xspace}

\newcommand{\reffig}[1]{Fig. \ref{#1}}

\newcommand{\refsec}[1]{Section \ref{#1}}
\newcommand{\etal}{\textit{et al.}\xspace}

\newcommand{\dd}{\,\mathrm{d}}

\DeclareMathOperator{\sign}{sign}
\DeclarePairedDelimiter\abs{\lvert}{\rvert}%

\renewcommand{\deg}{^{\circ}}
\newcommand{\norm}[1]{\left\lVert #1 \right\rVert}

\renewcommand{\vec}[1]{\mathbf{#1}}

\newcommand{\realnumbers}{\mathbb{R}}
\newcommand{\expect}{\mathbb{E}}

\algnewcommand\algorithmicforeach{\textbf{for each}}
\algdef{S}[FOR]{ForEach}[1]{\algorithmicforeach\ #1\ \algorithmicdo}

\renewcommand\algorithmicdo{}
\makeatletter
\renewcommand{\ALG@beginalgorithmic}{\footnotesize}
\makeatother

\newlength{\tempdima}
\newcommand{\rowname}[1]
{\rotatebox{90}{\makebox[\tempdima][c]{#1}}}

\title{\LARGE \bf
   Voxblox: Incremental 3D Euclidean Signed Distance Fields for On-Board MAV Planning
}

\author{Helen Oleynikova, Zachary Taylor, Marius Fehr, Roland Siegwart, and Juan Nieto\\
Autonomous Systems Lab, ETH Z{\"u}rich%
}

\begin{document}

\maketitle
\thispagestyle{empty}
\pagestyle{empty}

\begin{abstract}
Micro Aerial Vehicles (MAVs) that operate in unstructured, unexplored environments require fast and flexible local planning, which can replan when new parts of the map are explored.
Trajectory optimization methods fulfill these needs, but require obstacle distance information, which can be given by Euclidean Signed Distance Fields (ESDFs).

We propose a method to incrementally build ESDFs from Truncated Signed Distance Fields (TSDFs), a common implicit surface representation used in computer graphics and vision.
TSDFs are fast to build and smooth out sensor noise over many observations, and are designed to produce surface meshes.
Meshes allow human operators to get a better assessment of the robot's environment, and set high-level mission goals.

We show that we can build TSDFs faster than Octomaps, and that it is more accurate to build ESDFs out of TSDFs than occupancy maps.
Our complete system, called voxblox, will be available as open source and runs in real-time on a single CPU core.
We validate our approach on-board an MAV, by using our system with a trajectory optimization local planner, entirely on-board and in real-time.
\end{abstract}

\section{Introduction}
Rotary-wing Micro Aerial Vehicles (MAVs) have become one of the most popular robotics research platforms, as their agility and small size makes them ideal for many inspection and exploration applications.
However, their low payload and power budget, combined with fast dynamics, requires fast and light-weight algorithms.
Planning in unstructured, unexplored environments poses a particularly difficult problem, as both mapping and planning have to be done in real-time.
In this work, we focus specifically on providing a map for local planning, which quickly finds feasible paths through changing or newly-explored environments.
Furthermore, humans often supervise high-level mission goals, and therefore we also aim to provide a human-readable representation of the environment.

While many algorithms are well-suited for global MAV planning, (such as RRTs, graph search methods, and mixed-integer convex programs), local re-planning requires algorithms that can find feasible (though not necessarily optimal) paths in minimal time.
Trajectory optimization-based planning methods are well suited to these problems, as they are very fast and able to deal with complex environments.
However, they require the distances to obstacles to be known at all points in a map, as well as distance gradients \cite{ratliff2009chomp, oleynikova2016continuous-time}.
These distance maps are usually computed from an occupancy map, most often in batch, but more recently some incremental approaches have appeared~\cite{lau2010improved}.
The main drawback of these methods is that the maximum size of the map must be known \textit{a priori}, and cannot be dynamically changed.
Additionally, occupancy maps such as Octomap~\cite{hornung2013octomap} are often difficult for a human operator to parse while attempting to operate the vehicle out of line of sight.

\begin{figure}[tb]
  \centering
  \includegraphics[width=1.0\columnwidth,trim=000 0 00 0 px, clip=true]{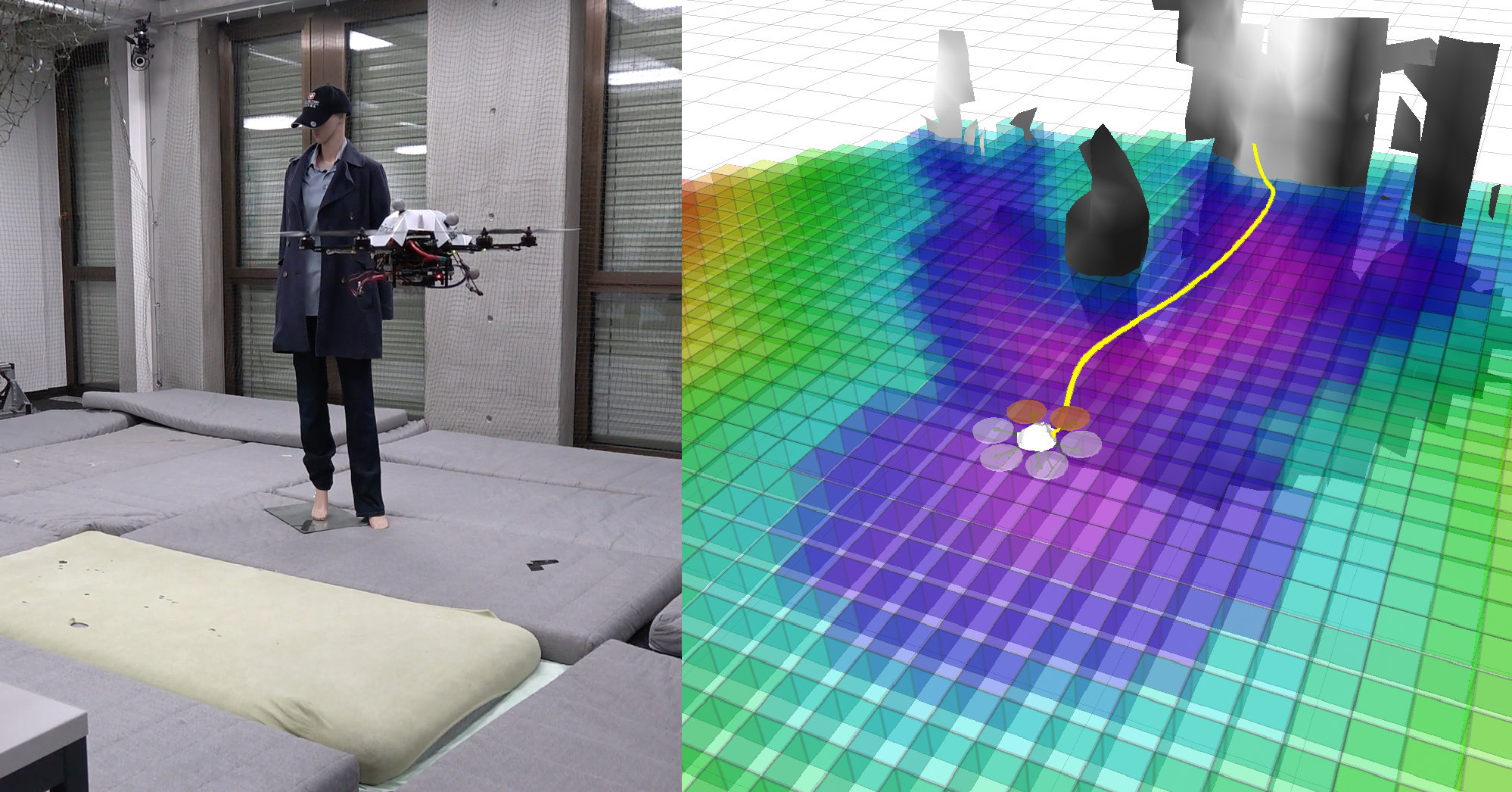}
  \caption{A planning experiment using voxblox-generated TSDF (shown as grayscale mesh) and ESDF (shown as a single horizontal slice of the 3D grid) running entirely in real-time and on-board the MAV. The vehicle attempts to plan to a point behind the mannequin by using a trajectory optimization-based method~\cite{oleynikova2016continuous-time} which relies on having smooth distance costs and gradients.}
  \label{fig:experiment}
\end{figure}

We attempt to overcome these shortcomings by proposing a system capable of incrementally building Euclidean Signed Distance Fields (ESDFs) online, in real-time on a dynamically growing map, while using an underlying map representation that is well-suited to visualization.
ESDFs are a voxel grid where every point contains its \textit{Euclidean} distance to the nearest obstacle.
Truncated Signed Distance Fields (TSDFs) have recently become a common implicit surface representation for computer graphics and vision applications \cite{curless1996volumetric, newcombe2011kinectfusion}, as they are fast to construct, filter out sensor noise, and can create human-readable meshes with sub-voxel resolution.
In contrast to ESDFs, they use \textit{projective} distance, which is the distance along the sensor ray to the measured surface, and calculate these distances only within a short \textit{truncation radius} around the surface boundary.
We propose to build ESDFs directly out of TSDFs and leverage the distance information already contained within the truncation radius, while also creating meshes for remote human operators.
We assume that the MAV is using stereo or RGB-D as the input to the map, and that its pose estimate is available.
Furthermore, we explore ways to optimize the construction of both TSDFs and ESDFs to run in real-time on a single CPU core, as MAVs are very computationally constrained and rarely have GPUs on board.

Our experiments on real datasets show that we can build TSDFs faster than Octomaps~\cite{hornung2013octomap}, which are commonly used for MAV planning.
We also analyze sources of error in our ESDF construction strategy, and validate the speed and accuracy of our method against simulated ground truth data.
Based on these results, we make recommendations on the best parameters for building both TSDFs and ESDFs for planning applications.
Finally, we show the complete system integrated and running in closed-loop as part of an online replanning strategy, entirely on-board an MAV.
This complete system, named \textbf{voxblox}, will be made available as an open-source library.

The contributions of this work are as follows:
\begin{easylist}[itemize]
  & Present the first method to incrementally build ESDFs out of TSDFs in dynamically growing maps.
  & Analyze different methods of building a TSDF to maximize reconstruction speed and surface accuracy at large voxel sizes. 
  & Provide both analytical and experimental analysis of errors in the final ESDF, and propose safety margins to overcome these errors. 
  & Validate the complete system by performing online replanning using these maps on-board an MAV. 
\end{easylist}

\section{Related Work}
\label{sec:related}
This section gives a brief overview of different map representations used for planning, and existing work in building ESDFs and TSDFs.

Occupancy maps are a common representation for planning.
One of the most popular 3D occupancy maps is called Octomap~\cite{hornung2013octomap}, which uses a hierarchical octree structure to store occupancy probabilities.
However, there are planning approaches for which only occupancy information is insufficient.
For example, trajectory optimization-based planners, such as CHOMP~\cite{zucker2013chomp}, require distances to obstacles and collision gradient information over the entire workspace of the robot.
This is usually obtained by building an ESDF in batch from another map representation.

While creating ESDFs or Euclidean Distance Transforms (EDTs) of 2D and 3D occupancy information is a well-studied problem especially in computer graphics, most recent work has focused on speeding up batch computations using GPUs \cite{cao2010parallel, teodoro2013efficient}.
However, our focus is to minimize computation cost on a CPU-only platform.

Lau \etal have presented an efficient method of incrementally building ESDFs out of occupancy maps~\cite{lau2010improved}.
Their method exploits the fact that sensors usually observe only a small section of the environment at a time, and significantly outperforms batch ESDF building strategies for robotic applications.
We extend their approach to be able to build ESDFs directly out of TSDFs, rather than from occupancy data, exploiting the existing distance information in the TSDF.

TSDFs, originally used as an implicit 3D volume representation for graphics, have become a popular tool in 3D reconstruction with KinectFusion~\cite{newcombe2011kinectfusion}, which uses the RGB-D data from a Kinect sensor and a GPU adaptation of Curless and Levoy's work~\cite{curless1996volumetric}, to create a system that can reconstruct small environments in real-time at millimeter resolution.

The main restriction of this approach is the fixed-size voxel grid, which requires a known map size and a large amount of memory.
There have been multiple extensions to overcome this shortcoming, including using a moving fixed-size TSDF volume and meshing voxels exiting this volume~\cite{whelan2012kintinuous}, using an octree-based voxel grid~\cite{steinbrucker2014volumetric}, and allocating blocks of fixed size on demand in a method called \textit{voxel hashing}~\cite{niessner2013real}.
We follow the voxel-hashing approach to allow our map to grow dynamically as the robot explores the environment.

The focus of all of these methods is to output a high-resolution mesh in real-time using marching cubes~\cite{lorensen1987marching}, frequently on GPUs.
There has also been work on speeding up these algorithms to run on CPU~\cite{steinbrucker2014volumetric} and even on mobile devices~\cite{klingensmith2015chisel}; however, the application of high-resolution 3D reconstruction remains the same.
Instead, our work focuses on creating representations that are accurate and fast enough to use for planning onboard mobile robots, while using large voxels to speed up computations and save memory.

One existing work that combines ESDFs and TSDFs is that of Wagner \etal, who use KinectFusion combined with CHOMP for planning for an armed robot~\cite{wagner20133d, wagner2013real}.
However, instead of updating the ESDF incrementally, they first build a complete TSDF, then convert it to an occupancy grid and compute the ESDF in a single batch operation for a fixed-size volume.
In contrast, our incremental approach gives us the ability to maintain an ESDF directly from a TSDF, handle dynamically growing the map without knowing its size \textit{a priori}, and is significantly faster than batch methods.

Features such as CPU computation time, incremental ESDF construction, and dynamically-growing map are essential for a map representation to use for on-board local planning for an MAV.

\section{System}
\label{sec:system}
Our overall system functions in two parts: first, incorporating incoming sensor data into a TSDF (described in detail in \refsec{sec:tsdf}), and then propagating updated voxels from the TSDF to update the ESDF (see \refsec{sec:esdf}).

In order to make it suitable for exploration and mapping applications, we use a dynamically sized map that makes use of the voxel hashing approach of Niessner \etal~\cite{niessner2013real}.
Each type of voxel (TSDF or ESDF) has its own layer, and each layer contains independent blocks that are indexed by their position in the map.
A mapping between the block positions and their locations in memory is stored in a hash table, allowing $\mathcal{O}(1)$ insertions and look-ups.
This makes the data structure flexible to growing maps, and additionally allows faster access than octree structures such as used by Octomap (which is $\mathcal{O}(\log n)$).
\begin{figure}[tb]
  \centering
  \includegraphics[width=1.0\columnwidth,trim=0 0 0 0 mm, clip=true]{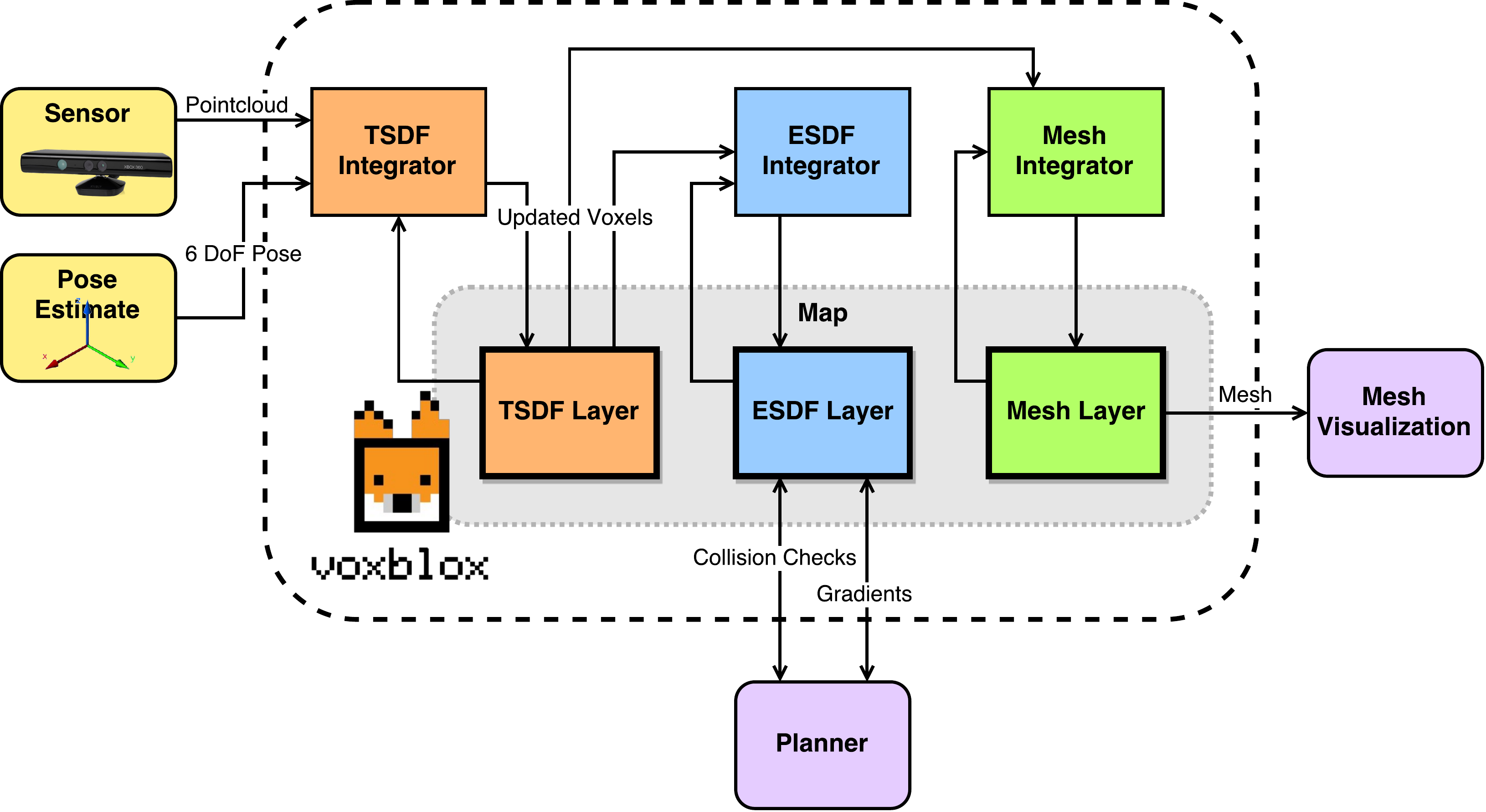}
  \caption{System diagram for voxblox, showing how the multiple map layers (TSDF, ESDF, and mesh) interact with each other and with incoming sensor data through integrators.}
  \label{fig:voxblox_system}
\end{figure}

\reffig{fig:voxblox_system} shows the overall system diagram.
The ESDF voxels and mesh may be reconstructed from updated TSDF voxels at any time in an incremental process, allowing on-demand ESDF generation for planning, or on-demand meshing for visualization.
The C++ mapping library implementing these algorithms, called \textbf{voxblox}, will be available as open-source.

\section{TSDF Construction}
\label{sec:tsdf}
Choices in how to build a TSDF out of sensor data can have a large impact on both the integration speed and the accuracy of the resulting reconstruction.
Here we present weighting and merging strategies, which increase accuracy and speed especially at large voxel sizes.

\subsection{Weighting}
\label{sec:weighting}
A common strategy to integrate a new scan into a TSDF is to ray-cast from the sensor origin to every point in the sensor data, and update the distance and weight estimates along this ray.
The choice of weighting function can have a strong impact on the accuracy of the resulting reconstruction, especially for large voxels, where thousands of points may be merged into the same voxel \textit{per scan}.

KinectFusion discussed using weights based on $\theta$, the angle between the ray from the sensor origin and the normal of the surface, however then stated that a constant weight of $1$ was sufficient to get good results for their millimeter-sized voxels~\cite{newcombe2011kinectfusion}.
This is a common approach in other literature~\cite{whelan2012kintinuous, niessner2013real, bylow2013real, kahler2015very}.

The general equations governing the merging are based on the existing distance and weight values of a voxel, $D$ and $W$, and the new update values from a specific point observation in the sensor, $d$ and $w$, where $d$ is the distance from the \textit{surface boundary}.
Given that $\vec{x}$ is the center position of the current voxel, $\vec{p}$ is the position of a 3D point in the incoming sensor data, $\vec{s}$ is the sensor origin,
and $\vec{x}, \vec{p}, \vec{s} \in \realnumbers^3$, the updated $D$ distance and $W$ weight values of a voxel at $\vec{x}$ will be:

\begin{eqnarray}
d(\vec{x}, \vec{p}, \vec{s}) & = & \norm{\vec{p} - \vec{x}} \sign \big((\vec{p} - \vec{x}) \bullet (\vec{p} - \vec{s})\big)\\
w_\textrm{const}(\vec{x}, \vec{p}) & = & 1 \label{eq:const_weight}\\
D_{i+1}(\vec{x}, \vec{p})& = & \frac{W_i(\vec{x}) D_i(\vec{x}) + w(\vec{x}, \vec{p}) d(\vec{x}, \vec{p})} {W_i(\vec{x}) + w(\vec{x}, \vec{p})} \label{eq:dist_merge}\\
W_{i+1}(\vec{x}, \vec{p}) & = & \min \big(W_i(\vec{x}) + w(\vec{x}, \vec{p}), W_{\textrm{max}}\big) \label{eq:weight_merge}
\end{eqnarray}

We propose a more sophisticated weight to compare to the constant weighting shown above.
Bylow \etal compared the effects of dropping the weight off behind the isosurface boundary, and found that a linear drop-off often yielded the best results~\cite{bylow2013real}.
Nguyen \etal empirically determined the RGB-D sensor model, and found that the $\sigma$ of a single ray measurement varied predominantly with $z^2$~\cite{nguyen2012modeling}, where $z$ is the depth of the measurement in the camera frame.
We combined a simplified approximation of the RGB-D model with the behind-surface drop-off as follows:
\begin{equation}
w_\textrm{quad}(\vec{x, p}) =\begin{cases}
\frac{1}{z^2} & -\epsilon < d  \\
\frac{1}{z^2} \frac{1}{\delta - \epsilon} (d + \delta) & -\delta < d < -\epsilon \\
0 & \phantom{-\epsilon <{}} d < -\delta, 
\end{cases}
\label{eq:quad_weight}
\end{equation}
where we use a truncation distance of $\delta = 4v$ and $\epsilon = v$, and $v$ is the voxel size.

Intuitively, this would make an even bigger difference in the presence of thin surfaces that are observed from multiple viewpoints, as this reduces the influence of voxels that have actually not been directly observed (those behind the surface).
An analysis of the effect this weighting has on surface reconstruction accuracy is presented in \refsec{sec:tsdf_results}.

\subsection{Merging}
\label{sec:merging}
As we focus on using larger voxel sizes to speed up computation, one important consideration is how new scans are merged into the existing voxel grid.
For large voxel sizes (on the order of tens of centimeters), thousands of rays from a single scan may map to the same voxel.
We exploit this for a significant speedup by designing a strategy that only performs raycasts once per end voxel.

There are two main methods for integrating information from a sensor data into a TSDF: raycasting~\cite{curless1996volumetric} and projection mapping~\cite{newcombe2011kinectfusion} \cite{klingensmith2015chisel}.

Raycasting casts a ray from the camera optical center to the center of each observed point, and updates all voxels from the center to truncation distance $\delta$ behind the point.
Projection mapping instead projects voxels in the visual field-of-view into the depth image, and computes its distance from the distance between the voxel center and the depth value in the image.
It is significantly faster, but leads to strong aliasing effects for larger voxels~\cite{klingensmith2015chisel}.

Our approach, \textit{grouped raycasting}, significantly speeds up raycasting without losing much accuracy.
For each point in the sensor scan, we project its position to the voxel grid, and group it with all other points mapping to the same voxel.
Then we take the weighted mean of all points and colors within each voxel, and do raycasting only once on this mean position. 
Since all measurements are still taken into account, and their weights and distances are combined as usual, this leads to a very similar reconstruction result while being up to 20 times faster than the naive raycasting approach, as shown in \refsec{sec:tsdf_results}.

\section{Constructing ESDF from TSDF}
\label{sec:esdf}
In this section we discuss how to build an ESDF for planning out of a TSDF built from sensor data, and then analyze bounds on the errors introduced by our approximations.

\subsection{Construction}
We base our approach on the work of Lau \etal, who present a fast algorithm for dynamically updating ESDFs from occupacy maps~\cite{lau2010improved}.
We extend their method to take advantage of TSDFs as input data, and additionally allow the ESDF map to dynamically change size.
The complete method is shown in Algorithm \ref{alg:esdf}, where $v_\textrm{T}$ represents a voxel in the original TSDF map and $v_\textrm{E}$ is the co-located voxel in the ESDF map.

One of the key improvements we have made is to use the distance stored in the TSDF map, rather than computing the distance to the nearest occupied voxel.
In the original implementation, each voxel had an occupied or free status that the algorithm could not change.
Instead, we replace this concept with a \textit{fixed} band around the surface: ESDF voxels that take their values from their co-located TSDF voxels, and may not be modified.
The size of the fixed band is defined by TSDF voxels whose distances fulfill $|v_\textrm{T}.d| < \gamma$, where $\gamma$ is the radius of the band, analyzed further in \refsec{sec:esdf_analysis}.

The general algorithm is based on the idea of \textit{wavefronts}~-- waves that propagate from a start voxel to its neighbors (using 26-connectivity), updating their distances, and putting updated voxels into the wavefront queue to further propagate to their neighbors.
We use two wavefronts: raise and lower.
A voxel gets added to the raise queue when its new distance value from the TSDF is higher than the previous value stored in the ESDF voxel.
This means the voxel, and all its children, need to be invalidated.
The wavefront propagates until no voxels are left with parents that have been invalidated.

The lower wavefront starts when a new fixed voxel enters the map, or a previously observed voxel lowers its value.
The distances of neighboring voxels get updated based on neighbor voxels and their distances to the current voxel.
The wavefront ends when there are no voxels left whose distance could decrease from its neighbors.

Unlike Lau \etal~\cite{lau2010improved}, who intersperse the lower and raise wavefronts, we raise all voxels first, then lower all voxels to reduce bookkeeping.
Additionally, where they treat unknown voxels as occupied, we do not update unknown voxels.
For each voxel, we store the direction toward the parent, rather than the full index of the parent.
For quasi-Euclidean distance (shown in the algorithm), this parent direction is toward an adjacent voxel, while for Euclidean distance, it contains the full distance to the parent.
A full discussion of Euclidean versus quasi-Euclidean distance is offered in the section below.

Finally, since new voxels may enter the map at any time, each ESDF voxel keeps track of whether it has already been observed.
We then use this in line \ref{line:neighbors} of Algorithm \ref{alg:esdf} to do a crucial part of book-keeping for new voxels: adding all of their neighbors into the \textit{lower} queue, so that the new voxel will be updated to a valid value.

Our approach incorporates a bucketed priority queue to keep track of which voxels need updates, with a priority of $\abs{d}$.
In the results, we compare two different variants: a FIFO queue and a priority queue (where the voxel with the smallest absolute distance is updated first).

\begin{algorithm}[tb]
  \caption{Updating ESDF from TSDF}
  \label{alg:esdf}
  \begin{algorithmic}[1] 
    \Function{Propagate}{$\textrm{map}_{\textrm{ESDF}}$, $\textrm{map}_{\textrm{TSDF}}$}
    \ForEach {voxel $v_\textrm{T}$ \textbf{in} updated voxels in $\textrm{map}_{\textrm{TSDF}}$}
    \If {\Call{isFixed}{$v_\textrm{T}$}}
    \If {$v_\textrm{E}.d > v_\textrm{T}.d$ \textbf{ or not }$v_\textrm{E}.\textrm{observed}$}
    \State $v_\textrm{E}.\textrm{observed} \gets$ True 
    \State $v_\textrm{E}.d \gets v_\textrm{T}.d$
    \State \Call{Insert}{lower, $v_\textrm{E}$}
    \Else 
    \State $v_\textrm{E}.d \gets v_\textrm{T}.d$
    \State \Call{Insert}{raise, $v_\textrm{E}$}
    \State \Call{Insert}{lower, $v_\textrm{E}$}
    \EndIf
    \Else
    \If {$v_\textrm{E}.\textrm{fixed}$}
    \State $v_\textrm{E}.\textrm{observed} \gets$ True 
    \State $v_\textrm{E}.d \gets \sign(v_\textrm{T}.d)\cdot d_\textrm{max}$
    \State \Call{Insert}{raise, $v_\textrm{E}$}
    \ElsIf {\textbf{not} $v_\textrm{E}.\textrm{observed}$}
    \State $v_\textrm{E}.\textrm{observed} \gets$ True 
    \State $v_\textrm{E}.d \gets \sign(v_\textrm{T}.d)\cdot d_\textrm{max}$
    \State \Call{InsertNeighbors}{lower, $v_\textrm{E}$} \label{line:neighbors}
    \EndIf
    \EndIf
    \EndFor
    \State \Call{ProcessRaiseQueue}{raise}
    \State \Call{ProcessLowerQueue}{lower}
    \EndFunction
    \Function{isFixed}{$v_\textrm{E}$}
    \Return $-\gamma < v_\textrm{E}.d < \gamma$
    \EndFunction
    \Function{InsertNeighbors}{queue, $v_\textrm{E}$}
    \ForEach {neighbor of $v_\textrm{E}$}
    \State \Call{Insert}{queue, neighbor}
    \EndFor
    \EndFunction
    \Function{ProcessRaiseQueue}{raise}
    \While{raise $\neq \emptyset$}
    \State $v_\textrm{E} \gets $ \Call{pop}{raise}
    \State $v_\textrm{E}.d \gets \sign(v_\textrm{E}.d)\cdot d_\textrm{max}$
    \ForEach {neighbor of $v_\textrm{E}$}
    \If {$v_\textrm{E}$.direction(neighbor) $=$ neighbor.parent}
    \State \Call{Insert}{raise, neighbor}
    \Else
    \State \Call{Insert}{lower, neighbor}
    \EndIf
    \EndFor
    \EndWhile
    \EndFunction
    \Function{ProcessLowerQueue}{lower}
    \While{lower $\neq \emptyset$}
    \State $v_\textrm{E} \gets $ \Call{pop}{lower}
    \ForEach {neighbor of $v_\textrm{E}$ at distance $\textrm{dist}$}
    \If {neighbor.$d > 0$ \textbf{and} $v_\textrm{E}.d + \textrm{dist} <$ neighbor.$d$ }
    \State {neighbor.$d \gets v_\textrm{E}.d + \textrm{dist}$}
    \State {neighbor.parent $\gets -v_\textrm{E}$.direction(neighbor)}
    \State \Call{Insert}{lower, neighbor}
    \ElsIf {neighbor.$d < 0$ \textbf{and} $v_\textrm{E}.d - \textrm{dist} >$ neighbor.$d$ }
    \State {neighbor.$d \gets v_\textrm{E}.d - \textrm{dist}$}
    \State {neighbor.parent $\gets -v_\textrm{E}$.direction(neighbor)}
    \State \Call{Insert}{lower, neighbor}    
    \EndIf
    \EndFor
    \EndWhile
    \EndFunction
  \end{algorithmic}
\end{algorithm}

\subsection{Sources of Error in ESDF}
\label{sec:esdf_analysis}
When using maps for planning, it is essential to know what effect the method has on the error in the final distance computations.
In this section, we aim to quantify the effect of our approximations and recommend a safety margin by which to increase bounding boxes used for planning.

We consider two key contributions to error in the final ESDF: first, the TSDF projective distance calculations, and second, the quasi-Euclidean approximation in distance calculations.

Projective distance (distance along the camera ray to the surface) will always match or overestimate the actual Euclidean distance to the nearest surface.
Therefore, to use projective distances from the TSDF, we need to quantify the error this will introduce.
The error is dependent on $d$, the measured distance of the voxel, and $\theta$, the incidence angle between the camera ray and the object surface.
We assume locally planar objects.
The projective error residual $r_p(\theta)$ can therefore be expressed as:
\begin{equation}
r_p(\theta) = d \sin(\theta) - d
\end{equation}

For the purposes of this analysis, we assume that the incidence angle $\theta$ can be between $\pi/20$ and $\pi/2$, and is uniformly distributed in this range.
The lower bound of $\pi/20$ comes from the observation that a camera ray can not be parallel to a surface boundary, nor can a camera exist infinitesimally close to a surface, due to the physical dimensions of the camera. $\pi/20$ corresponds to an MAV a minimum of 1 meter away from a surface with a maximum sensor ray length of 5 meters.
Given that $f(\theta)$ is the uniform distribution between $\pi/20$ and $\pi/2$, as this is symmetric, then the expected error for a single voxel observation will be:
\begin{align}
\expect[r_p(\theta)] &  =  \int_{\frac{\pi}{20}}^{\frac{\pi}{2}} \frac{20}{9 \pi} (d \sin(\theta) - d)\dd \theta \nonumber \\
&  =  -0.3014 d 
\end{align}
Note that $d$ has an upper bound of the truncation distance $\delta$. 

\begin{figure}[tb]
  \centering
  \begin{subfigure}[b]{0.51\columnwidth}
    \includegraphics[width=1.0\columnwidth,trim=0 00 0 00 mm, clip=true]{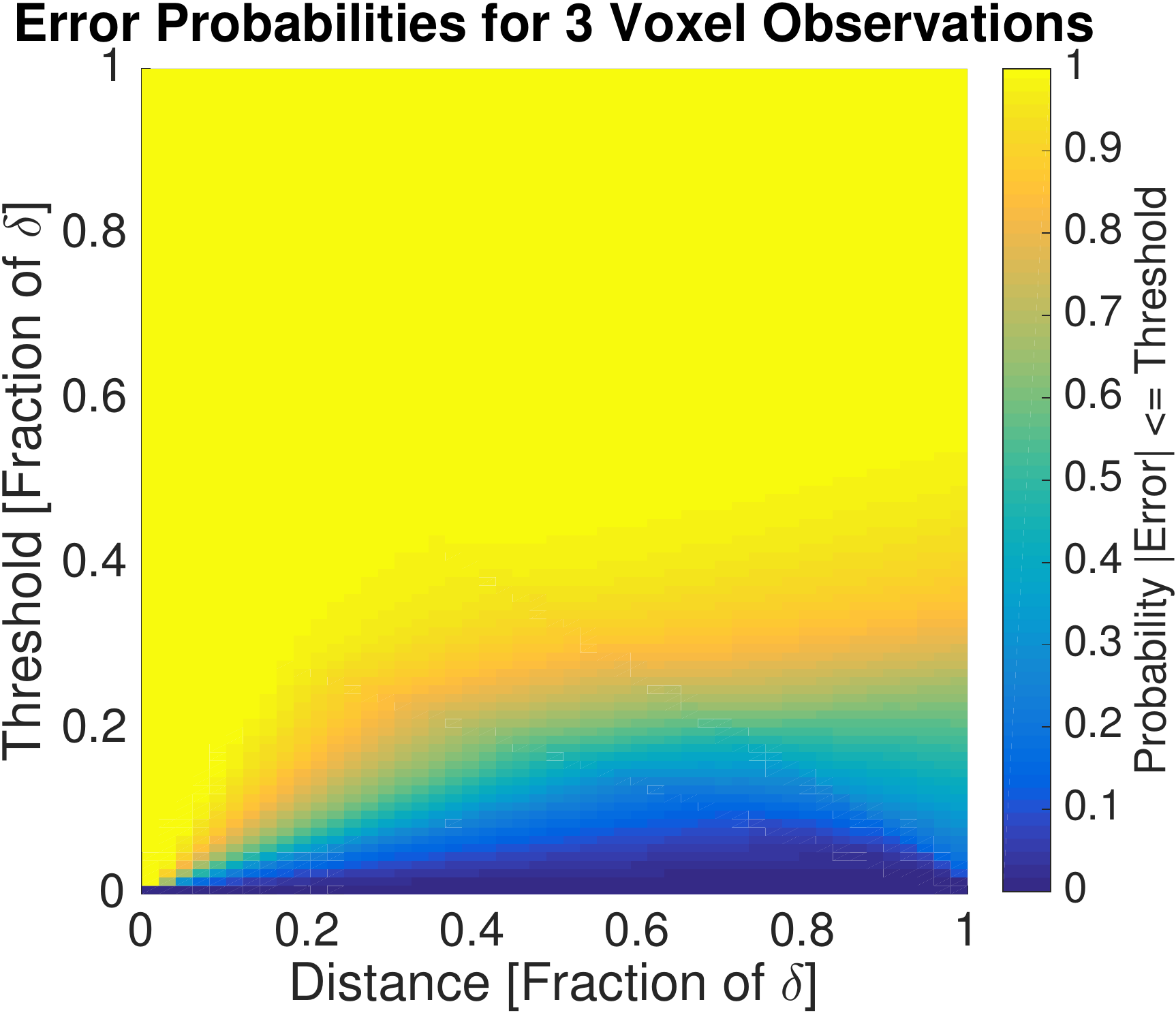}
    \caption{}
    \label{fig:mc_3_obs}
  \end{subfigure}
  \begin{subfigure}[b]{0.46\columnwidth}
    \includegraphics[width=1.0\columnwidth,trim=0 0 0 0 mm, clip=true]{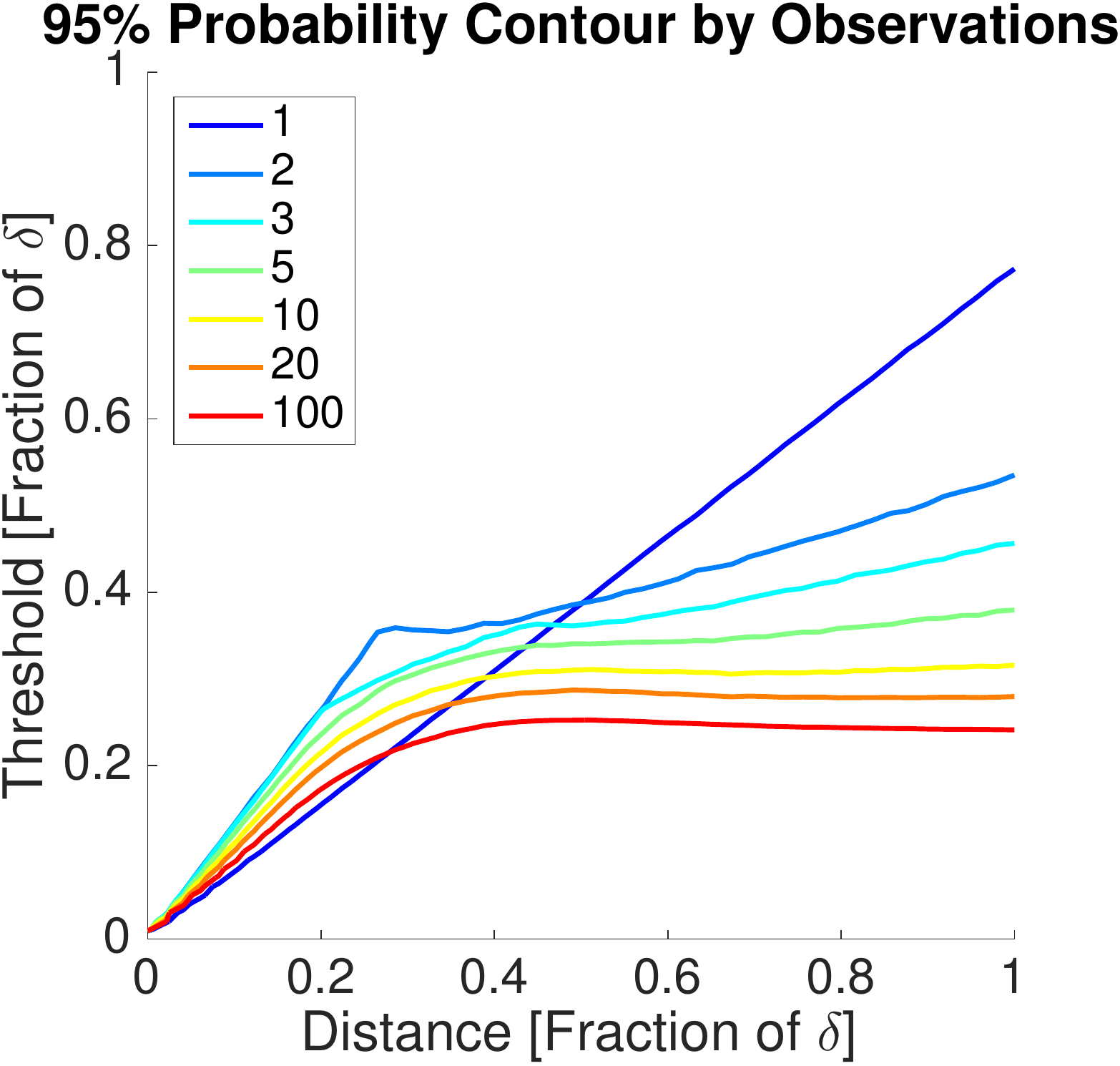}
    \caption{}
    \label{fig:mc_mult_obs}
  \end{subfigure}
  \caption{
    Probability of the distance error being below a threshold, for a given voxel distance measurement. \subref{fig:mc_3_obs} shows the probabilities for 3 voxel observations, and \subref{fig:mc_mult_obs} shows $95\%$ probability contours for multiple observations. For a single voxel observation, the maximum error with $p=0.95$ is $0.8\delta$, while for 3 observations $p=0.95$ falls at $0.5\delta$, and trends towards $0.25\delta$ as the number of observations grows.}
  \label{fig:monte_carlo}
\end{figure}

However, this does not consider multiple observations of the same voxel, which will lower this error.
To quantify this, we performed Monte Carlo simulations of merging multiple independent measurements of the same voxel, shown in \reffig{fig:monte_carlo}.
The results show that even for as few as 3 observations, the error has an upper bound of $0.5\delta$ with $p = 0.95$, and as the number of observations increases, the error at this probability is reduced down to below $0.25\delta$.

Depending on how large the fixed band is determines how much to compensate for this error. If only a single voxel of the surface frontier is used, then it is safe to increase the safety distance by half of one voxel.

The second source of error considered is from the quasi-Euclidean distance assumption in the ESDF calculations.
Quasi-Euclidean distance is measured along horizontal, vertical, and diagonal lines in the grid, leading to no error when the angle $\phi$ between the surface normal and the ray from the surface to the voxel is a multiple of $45 \deg$, and a maximum error at $\phi = 22.5 \deg$~\cite{montanari1968method}.
If $\phi$ is uniformly distributed between $0$ and $\pi/4$ the residual $r_q(\phi)$ for this error, and its maximum and expected values are:
\begin{align}
r_q(\phi) & = \Big(d - \frac{d \sin(5\pi/8 - \phi)}{\sin(3\pi/8)} \Big) \\
r_q(\frac{\pi}{8}) & = -0.0824d \\
\expect[r_q(\phi)] & =  \int_{0}^{\frac{\pi}{4}} \frac{4}{\pi} \Big(d - \frac{d \sin(5\pi/8 - \phi)}{\sin(3\pi/8)} \Big) \dd\phi \nonumber \\
& =  -0.0548 d
\end{align}
Since the $d$ in this case only has an upper bound in the maximum ESDF computed distance, we recommend inflating the bounding box of the robot by $8.25\%$.
\refsec{sec:esdf_results} has empirical results on what effect this assumption has on the overall error in the ESDF computations, and shows that in practice it is small enough to justify the speed-up between full Euclidean and quasi-Euclidean distance.

\section{Experimental Results}
\label{sec:results}
In this section we validate the algorithms presented above on two real-world datasets: the cow dataset with an RGB-D sensor and EuRoC with a stereo camera, both validated against structure ground truth.

The cow dataset features several objects including a large fiberglass cow in a small room.
It is taken with the original Microsoft Kinect, uses pose data from a Vicon motion capture system, and the ground truth is from a Leica TPS MS50 laser scanner with 3 scans merged together.
It will made publicly available with this publication.

The EuRoC dataset is a public benchmark on 3D reconstruction accuracy~\cite{burri2016euroc}, in a medium-sized room filled with objects.
It is taken with a narrow-baseline grayscale stereo sensor, using Vicon fused with IMU as pose information, and Leica TPS MS50 scans as structure ground truth.
We use the \texttt{V1\_01\_easy} dataset for experiments.

All experiments are done on a quad-core i7 CPU at 2.5 GHz. Only one thread is used.

\subsection{TSDF Construction}
\label{sec:tsdf_results}

\begin{figure}
  \centering
  \settoheight{\tempdima}{\includegraphics[width=.30\columnwidth]{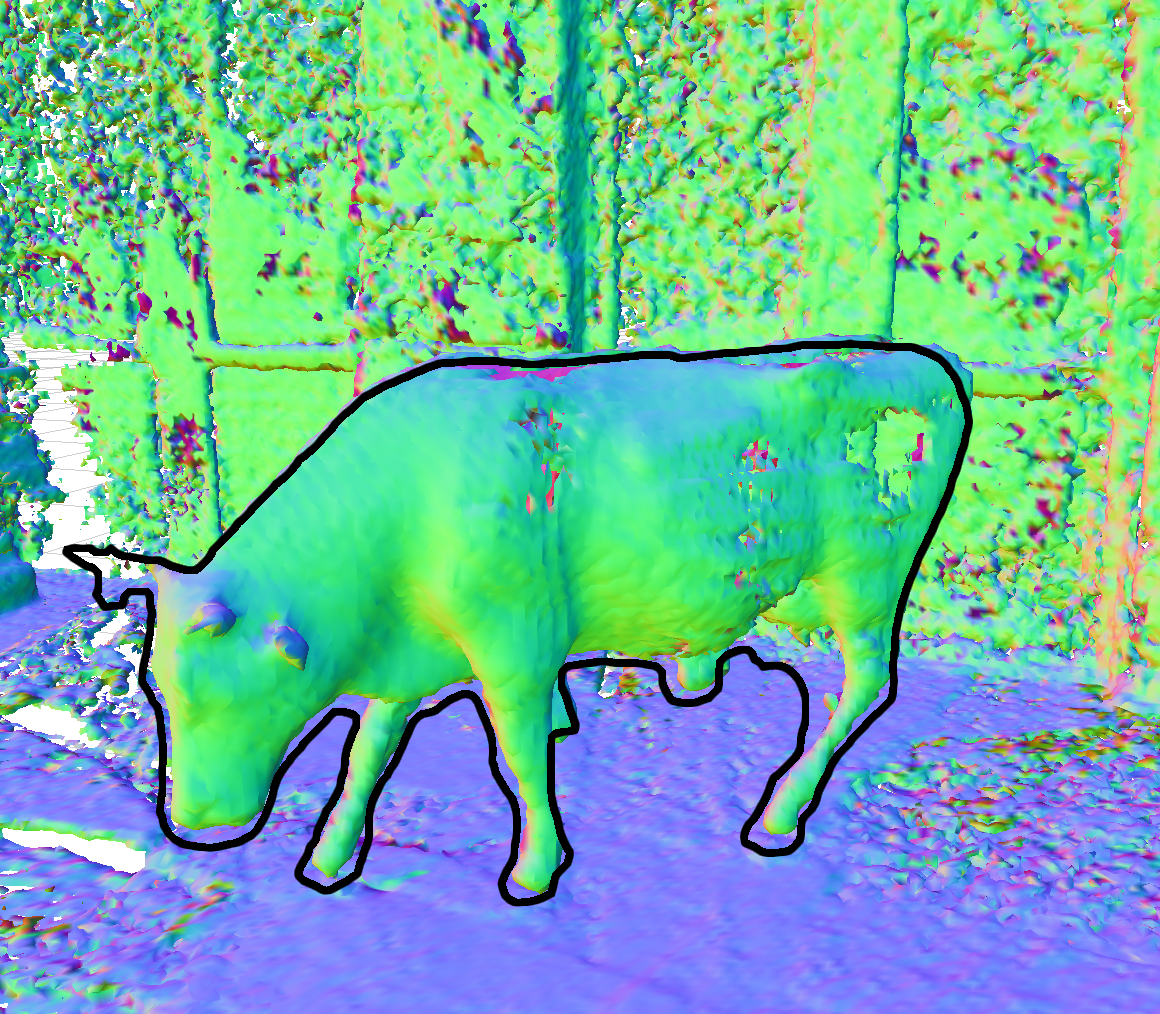}}%
  \centering\begin{tabular}{@{}c@{ }c@{ }c@{ }c@{}}
    \rowname{Constant}&
    \includegraphics[width=.30\columnwidth,trim=000 000 100 200 mm, clip=true]{figures/cow_results/overlay_const_0_02.png}&
    \includegraphics[width=.30\columnwidth,trim=000 000 100 200 mm, clip=true]{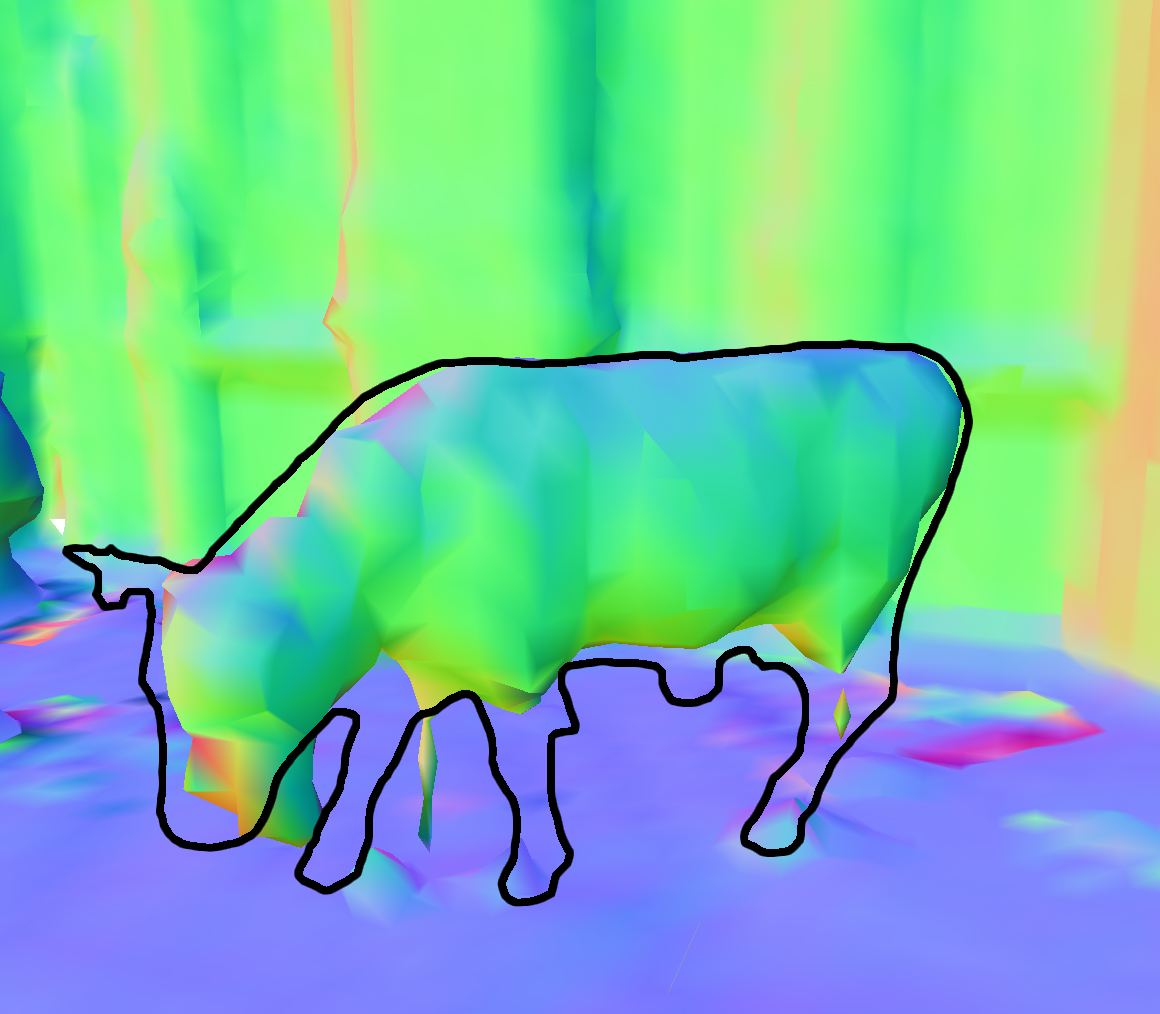}&
    \includegraphics[width=.30\columnwidth,trim=000 000 100 200 mm, clip=true]{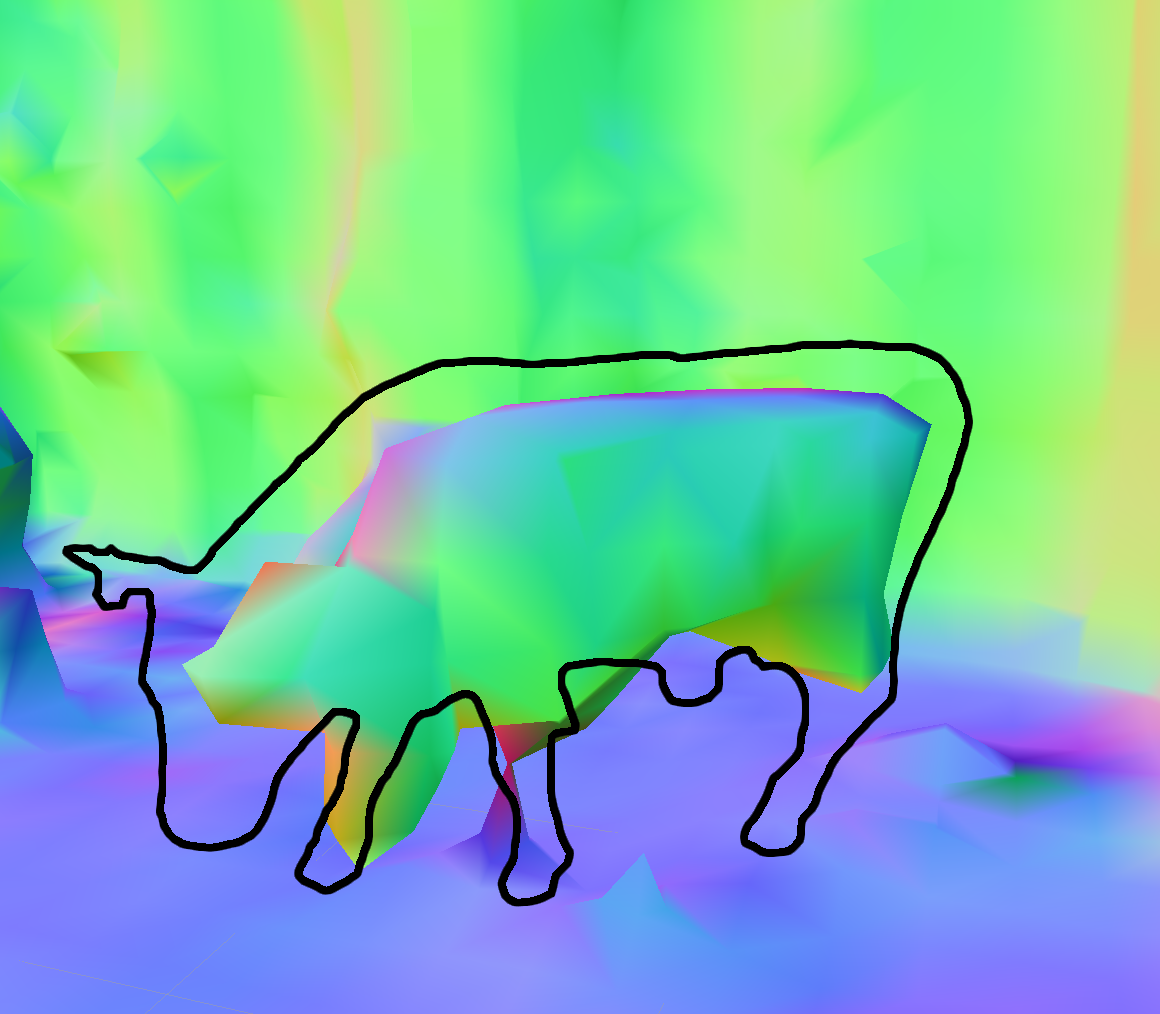}\\[-1.5ex]
    \rowname{Our Weight}&
    \includegraphics[width=.30\columnwidth,trim=000 000 100 200 mm, clip=true]{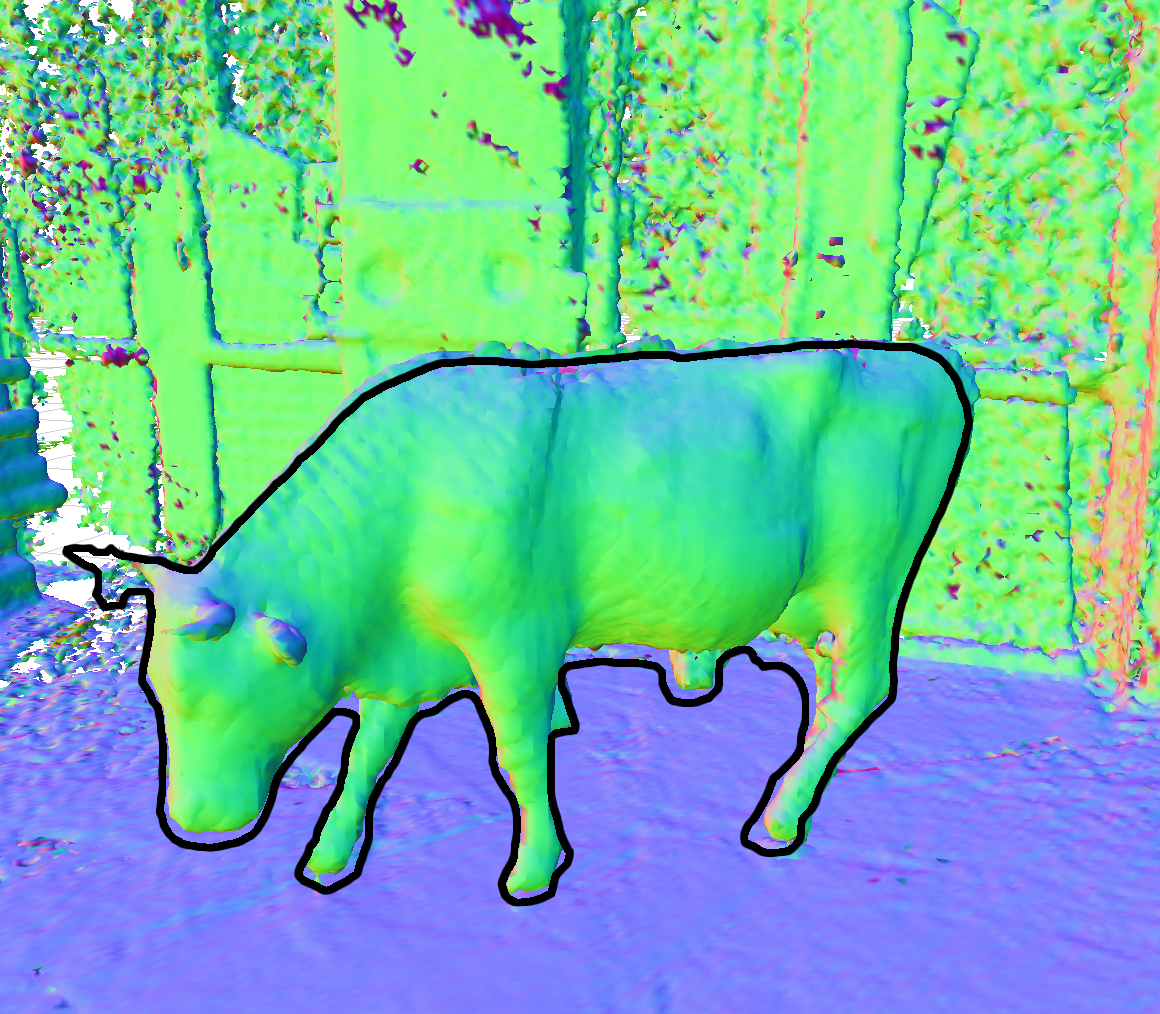}&
    \includegraphics[width=.30\columnwidth,trim=000 000 100 200 mm, clip=true]{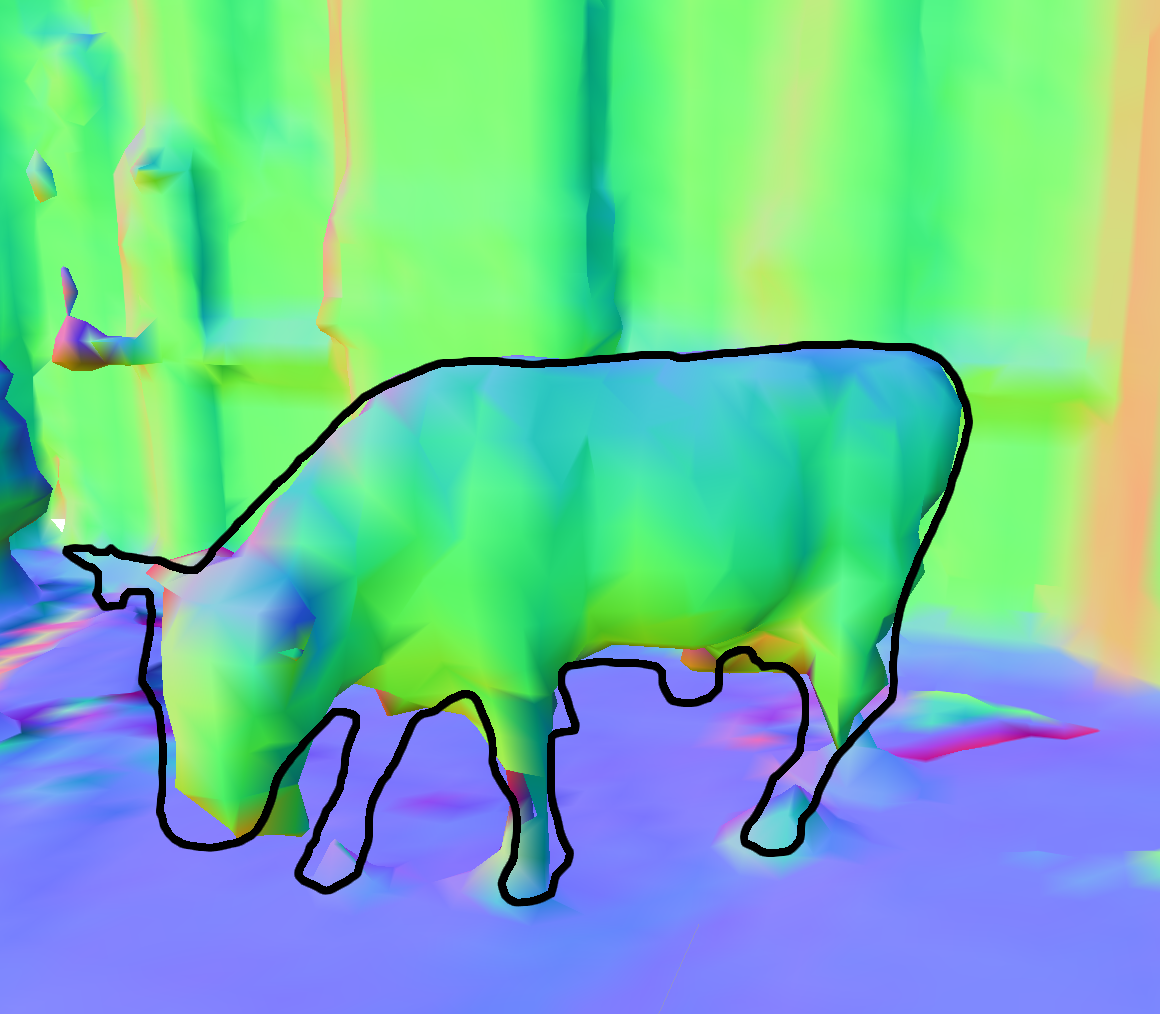}&
    \includegraphics[width=.30\columnwidth,trim=000 000 100 200 mm, clip=true]{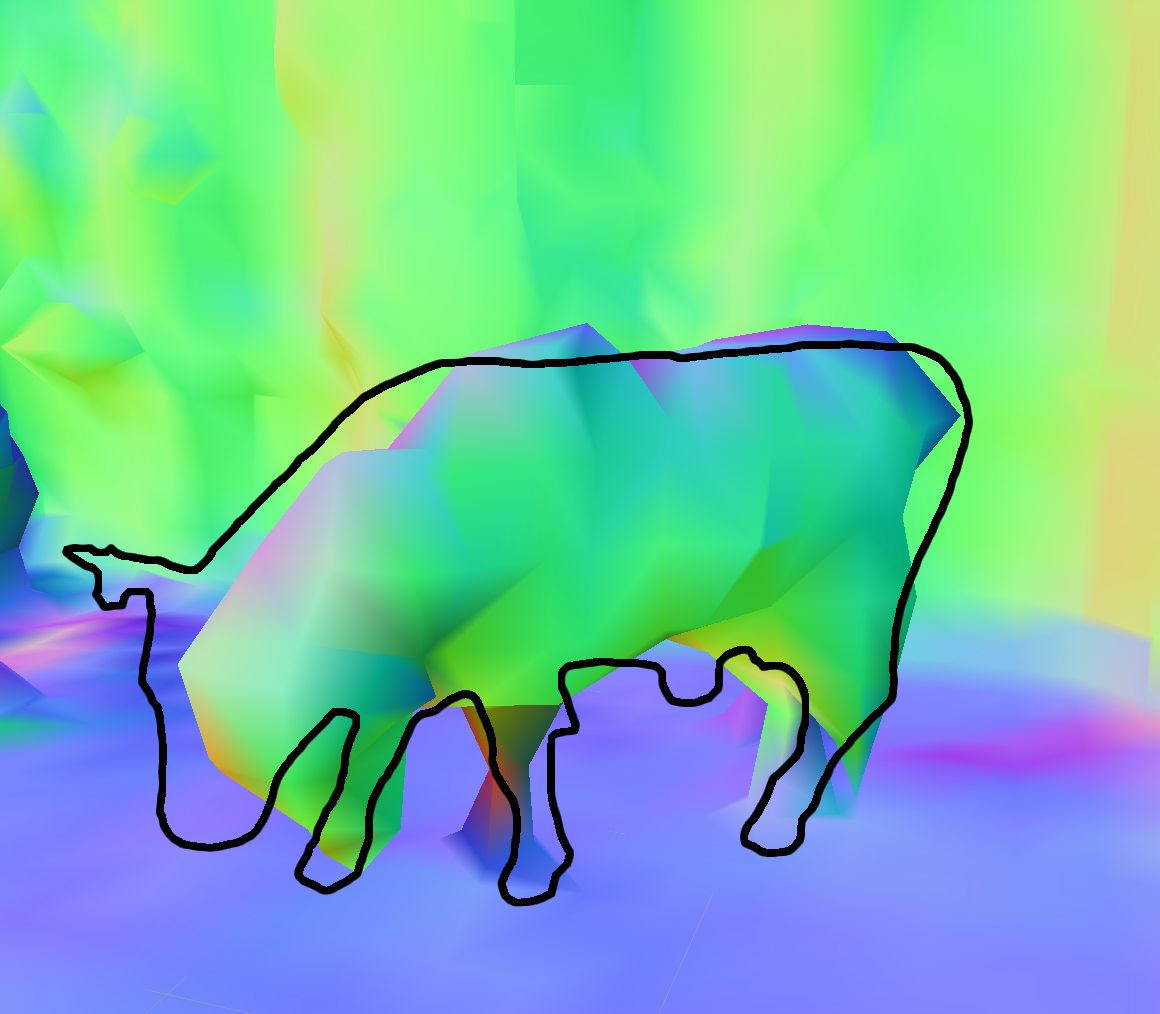}\\[-0.5ex]
    &0.02 m & 0.10 m & 0.20 m 
  \end{tabular}
  \caption{Qualitative comparisons of weighting/merging strategies on the cow dataset, colored by normals and with the object outline from ground truth overlaid. As can be seen, especially at large voxel sizes, our weighting strategy distorts the structure less.}
  \label{fig:cow_compare}
\end{figure}

\begin{figure}[tb]
  \centering
  \includegraphics[width=0.6\columnwidth,trim=0 0 0 0 px, clip=true]{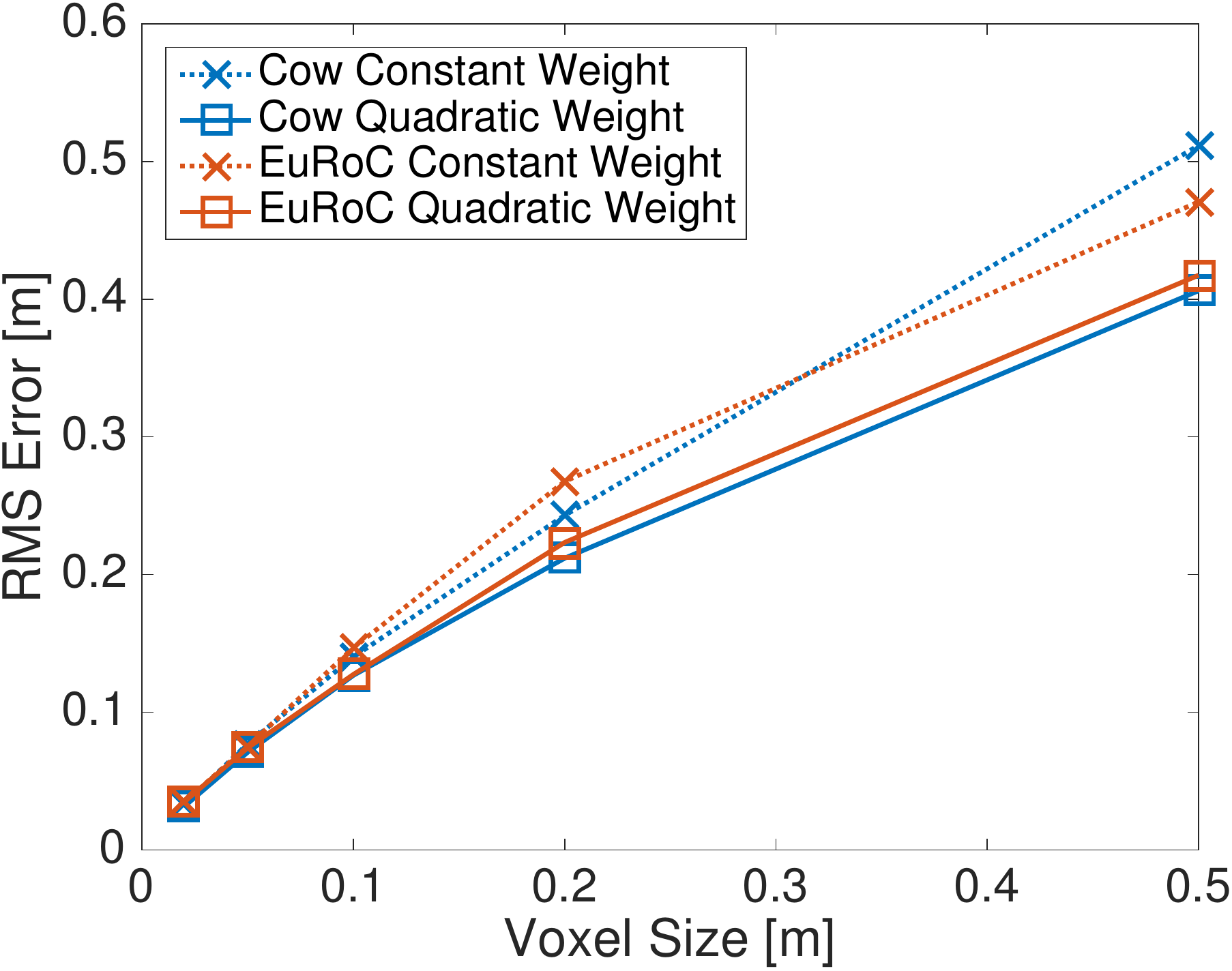}
  \caption{Structure accuracy reconstruction results for TSDFs with various voxel sizes, and comparing constant weight and quadratic weight with linear drop-off behing the surface. It can be seen that the choice of weighting function makes a more significant difference at larger voxel sizes, as more measurements are combined into any given voxel.}
  \label{fig:tsdf_results}
\end{figure}

\begin{figure}[tb]
  \centering
  \includegraphics[width=0.6\columnwidth,trim=0 0 0 0 mm, clip=true]{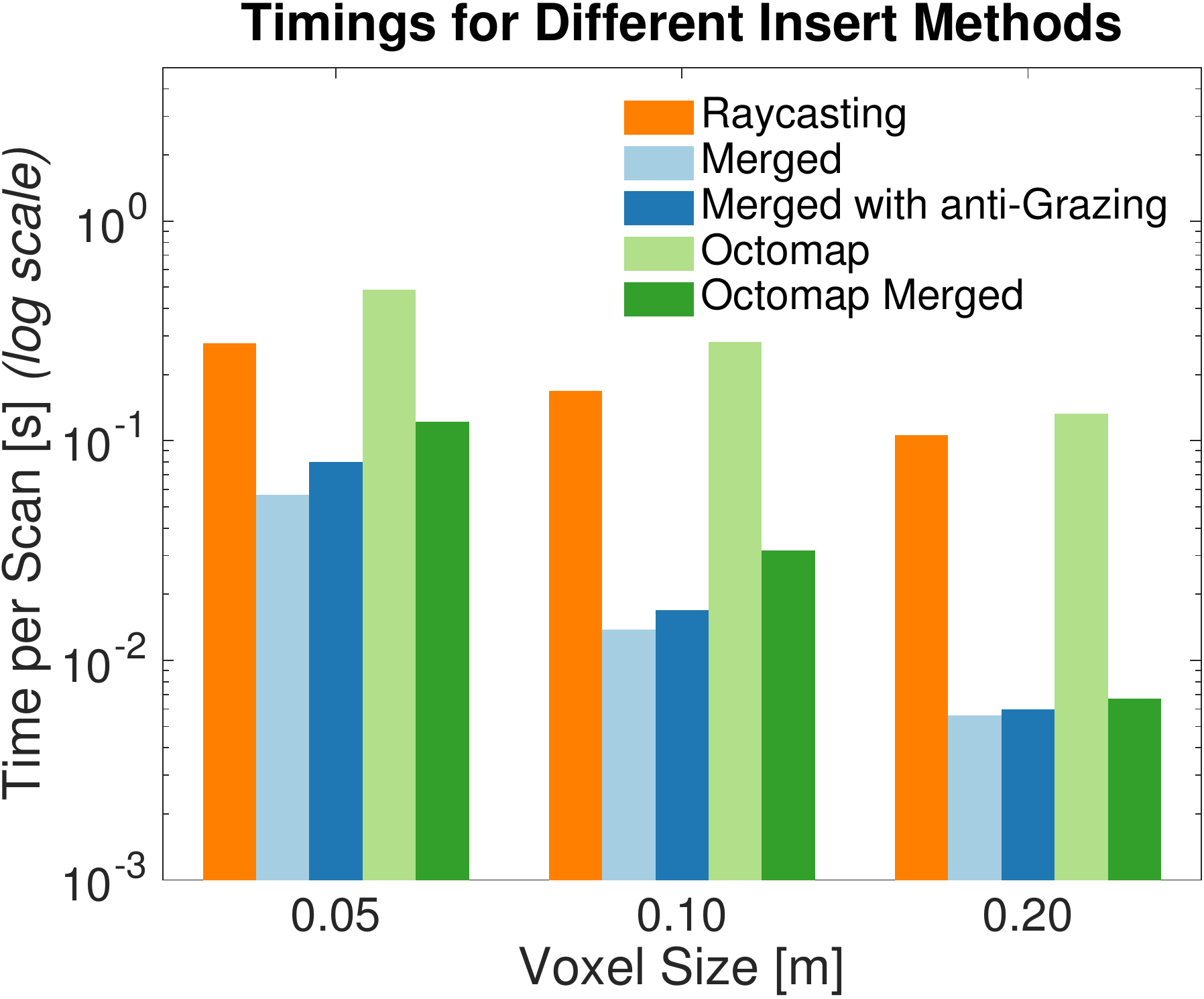}
  \caption{Timing results for different merging strategies on the EuRoC dataset. Our approach is up to 20 times faster than standard raycasting into a TSDF, and up to 2 times faster than even grouped Octomap insertions. Note log time scale.}
  \label{fig:timing}
\end{figure}

In order to verify that our weighting strategy scales well with larger voxel sizes, we validate our TSDF reconstructions against the structure ground truth for our datasets.

We evaluate the accuracy of our reconstruction by projecting each point in the ground truth pointcloud into the TSDF, performing trilinear interpolation to get the best estimate of the distance at that point, and taking that distance as an error.
We consider only known voxels, and allow a maximum error equal to the truncation distance ($\delta = 4v$).

Qualitative comparison are shown on the cow dataset in \reffig{fig:cow_compare}, compared to the ground truth cow silhouette. 
As can be seen, constant weighting significantly distorts the geometry of the cow at larger voxel sizes: the head is no longer in the correct position, and the rear legs are gone entirely, which will lead to incorrect distance estimates in the ESDF, while our weighting strategy better preserves structure.

\reffig{fig:tsdf_results} shows a quantitative comparison on both datasets with respect to voxel size: as can be seen, weighting has a more significant effect on error as voxel size increases, and our proposed quadratic weighting always outperforms constant weighting.

A comparison of the timings between various merging strategies and against Octomap~\cite{hornung2013octomap} is shown in \reffig{fig:timing}.
While Octomap with the grouped raycasting strategy as discussed in \refsec{sec:merging} is already significantly faster than normal raycasting Octomap, it is still substantially slower than our TSDF approach. 
This is due to the hierarchical data structure: as the number of nodes in the Octomap grows larger, look-ups in the tree get slower, as they scale with $\mathcal{O}(\log n)$; with voxel hashing~\cite{niessner2013real}, the lookups remain $\mathcal{O}(1)$.
Grouped raycasting leads to significant speeds up, especially with larger voxel sizes (as more points project into the same voxel).
Overall, we show that using our merging strategy makes using TSDFs feasible on a single CPU core, allowing it to be used for real-time mapping and planning applications on-board an MAV.

\subsection{ESDF Construction}
\label{sec:esdf_results}
\subsubsection{Simulation Results}
\begin{figure}[tb]
  \centering
  \includegraphics[width=0.5\columnwidth,trim=0 0 0 0 px, clip=true]{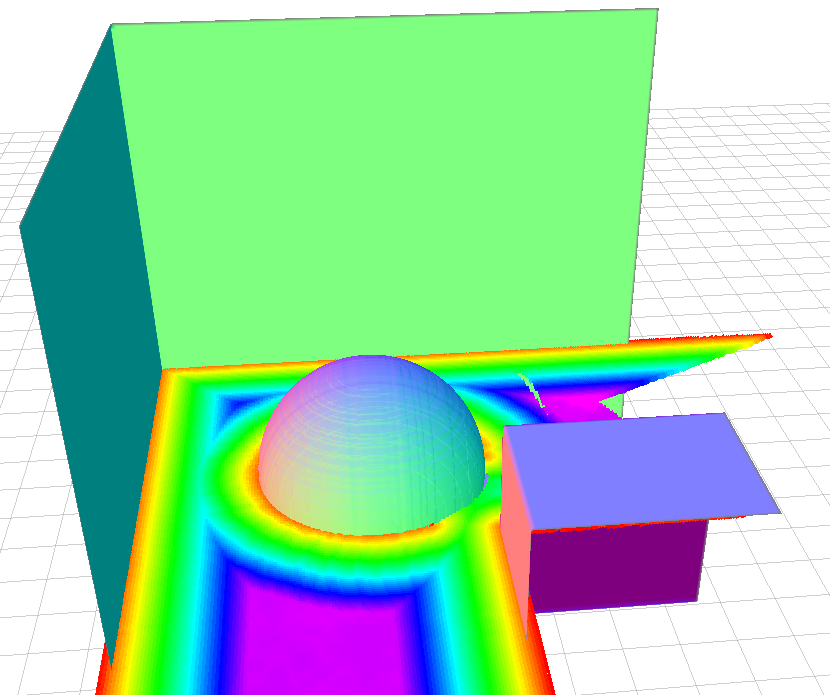}
  \caption{A normal-colored ground-truth mesh of the simulation experiment, with 3 planes (not pictured: ground plane), a cube, and a sphere. Also shown is a horizontal slice of the ESDF generated from 50 random viewpoints with a voxel size of 0.05 meters.}
  \label{fig:esdf_setup}
\end{figure}

To evaluate the errors introduced by various ESDF construction methods, we set up a simulated benchmark with 3 planes, a sphere, and a cube, shown with a horizontal ESDF slice in \reffig{fig:esdf_setup}, of size $10\times10\times10$ meters.
We simulated a noiseless RGB-D sensor with a resolution of $320\times240$ and a maximum distance of 5 meters. Readings were taken at 50 random free-space locations, uniformly sampled from all 6 DoF poses in the space that were a minimum of 1 meter from an obstacle.

\begin{figure}[tb]
  \centering
  \begin{subfigure}[b]{0.48\columnwidth}
    \includegraphics[width=1.0\columnwidth,trim=0 0 0 0 px, clip=true]{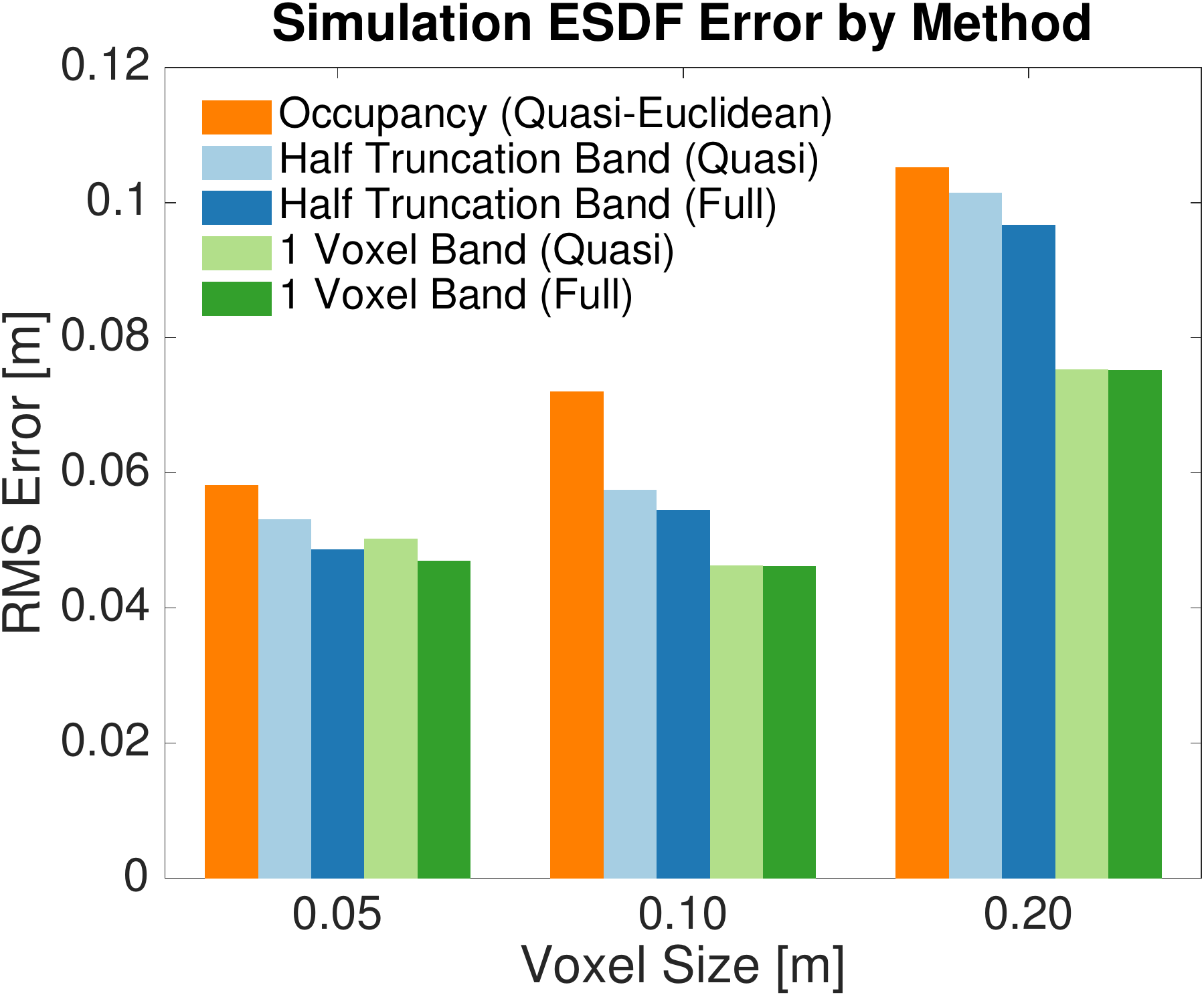}
    \caption{}
    \label{fig:esdf_sim_error}
  \end{subfigure}
  \begin{subfigure}[b]{0.48\columnwidth}
    \includegraphics[width=1.0\columnwidth,trim=0 0 0 0 mm, clip=true]{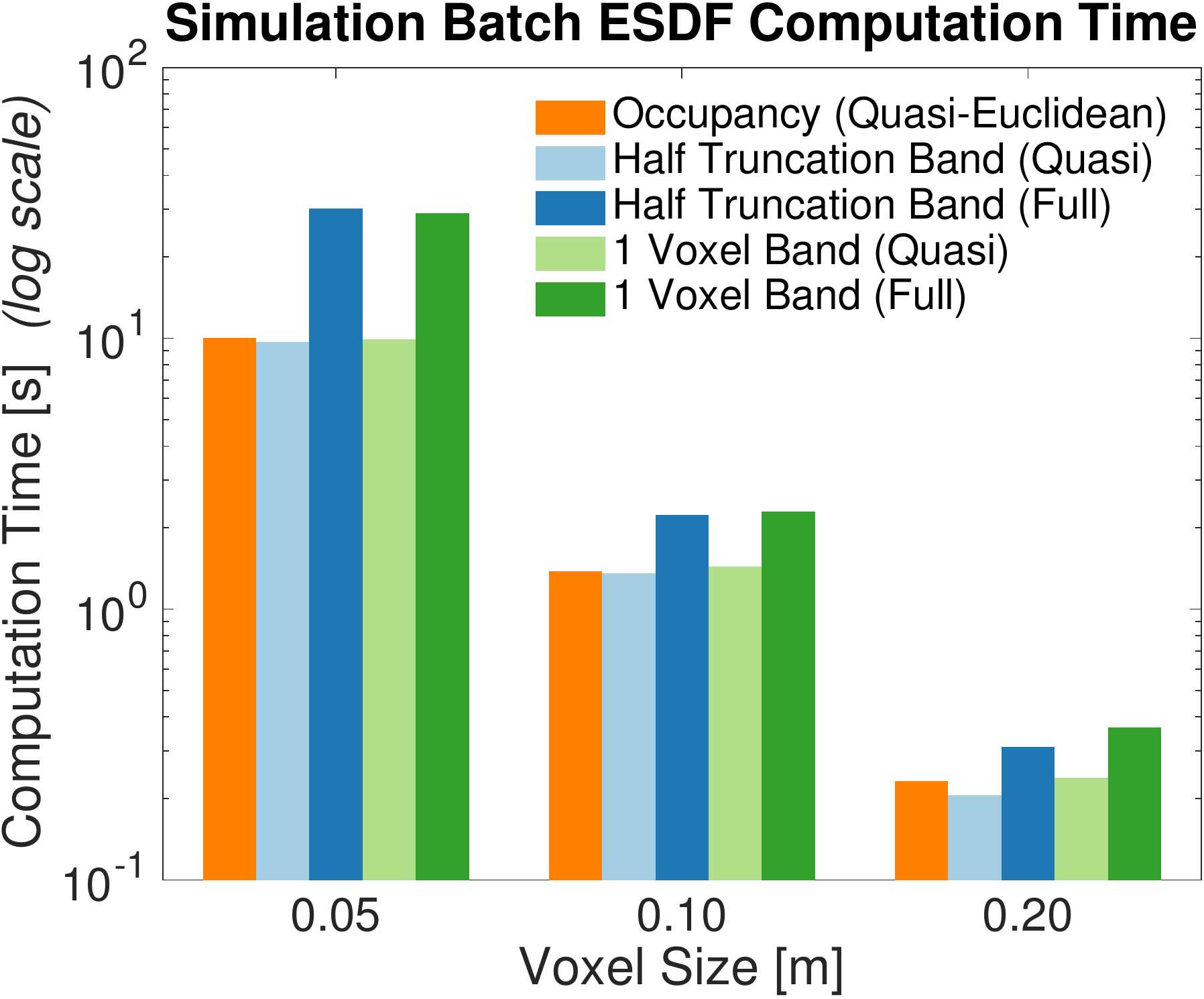}
    \caption{}
    \label{fig:esdf_sim_time}
  \end{subfigure}
  \caption{A comparison of methods for generating ESDFs from TSDFs and occupancy, and their errors and timing differences. Using all of the data in half the truncation distance is more accurate thanoccupancy, while using only one voxel-width of the surface outperforms all other methods, as shown in \subref{fig:esdf_sim_error}. Quasi-Euclidean distance has only a marginal increase in error for a large decrease in computation time \subref{fig:esdf_sim_time}.}
  \label{fig:sim_esdf_results}
\end{figure}

We produced ground truth ESDFs of the space by evaluating the minimum distance to the objects at voxel centers.
We then built a TSDF out of the simulated sensor data, and used multiple ESDF building methods to compare their error against this ground truth, as well as their integration times, shown in \reffig{fig:sim_esdf_results}.
The most basic method, occupancy, treats all negative-valued TSDF voxels as occupied and assigns them a distance of 0, similar to Wagner \etal ~\cite{wagner2013real}
The next set of methods takes a band of values around the surface of the TSDF of half the size of the truncation distance ($\delta/2 < |d|$), and we compare both full Euclidean and quasi-Euclidean distances.
The last set of methods takes a one-voxel-wide band around the surface, with both quasi-Euclidean and Euclidean distance.

As can be seen, the lowest errors are found by taking a one-voxel-wide fixed band around the surface.
This is due to the projection error discussed in \refsec{sec:esdf_analysis}.
However, it is important to note that all of the methods have a significantly lower error than using occupancy values, showing the advantages of building these maps out of TSDFs rather than occupancy maps.

While using full Euclidean distance shows improvement in ESDF error of 8.23\%, 5.18\%, and 4.72\% in the half-truncation distance method for voxel sizes of 0.05, 0.10, and 0.20 meters, respectively, the integration times are increased by 201.0\%, 61.3\%, and	33.9\%.
Given that real-time execution is one of the core goals of this approach, our findings show that for many applications, using quasi-Euclidean distance is a good trade-off between error and runtime.

\subsubsection{Real Data}
\begin{figure}[tb]
  \centering
  \includegraphics[width=0.6\columnwidth,trim=0 0 0 0 mm, clip=true]{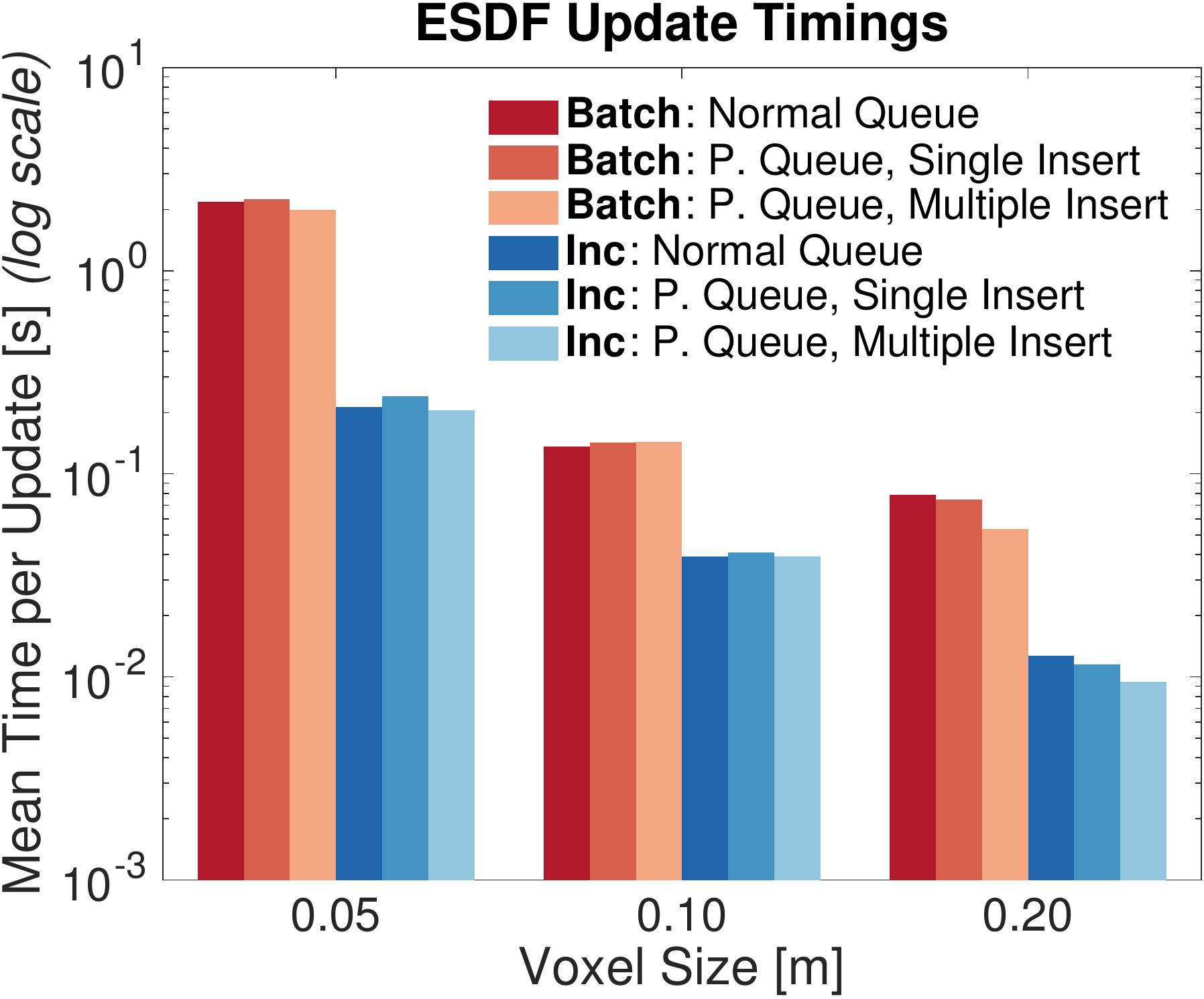}
  \caption{Timing results for updating ESDF in batch and incrementally, with different queueing strategies on the EuRoC dataset. The normal FIFO queue performs best for small voxel sizes, and at large voxel sizes, there is a speed-up from using a single-insert priority queue. Note the log time scale.}
  \label{fig:esdf_timing}
\end{figure}

To validate the presented ESDF runtimes, we used real data from the EuRoC dataset for our evaluations on incremental and batch timings, using two different queueing methods, as discussed in \refsec{sec:esdf}.
It can be seen in \reffig{fig:esdf_timing} that building the ESDF incrementally leads to an order of magnitude speedups over the entire dataset, and that at large voxel sizes, using a single-insert priority queue is faster than using a FIFO queue.

\begin{figure}[tb]
  \centering
  \includegraphics[width=0.6\columnwidth,trim=0 0 0 0 mm, clip=true]{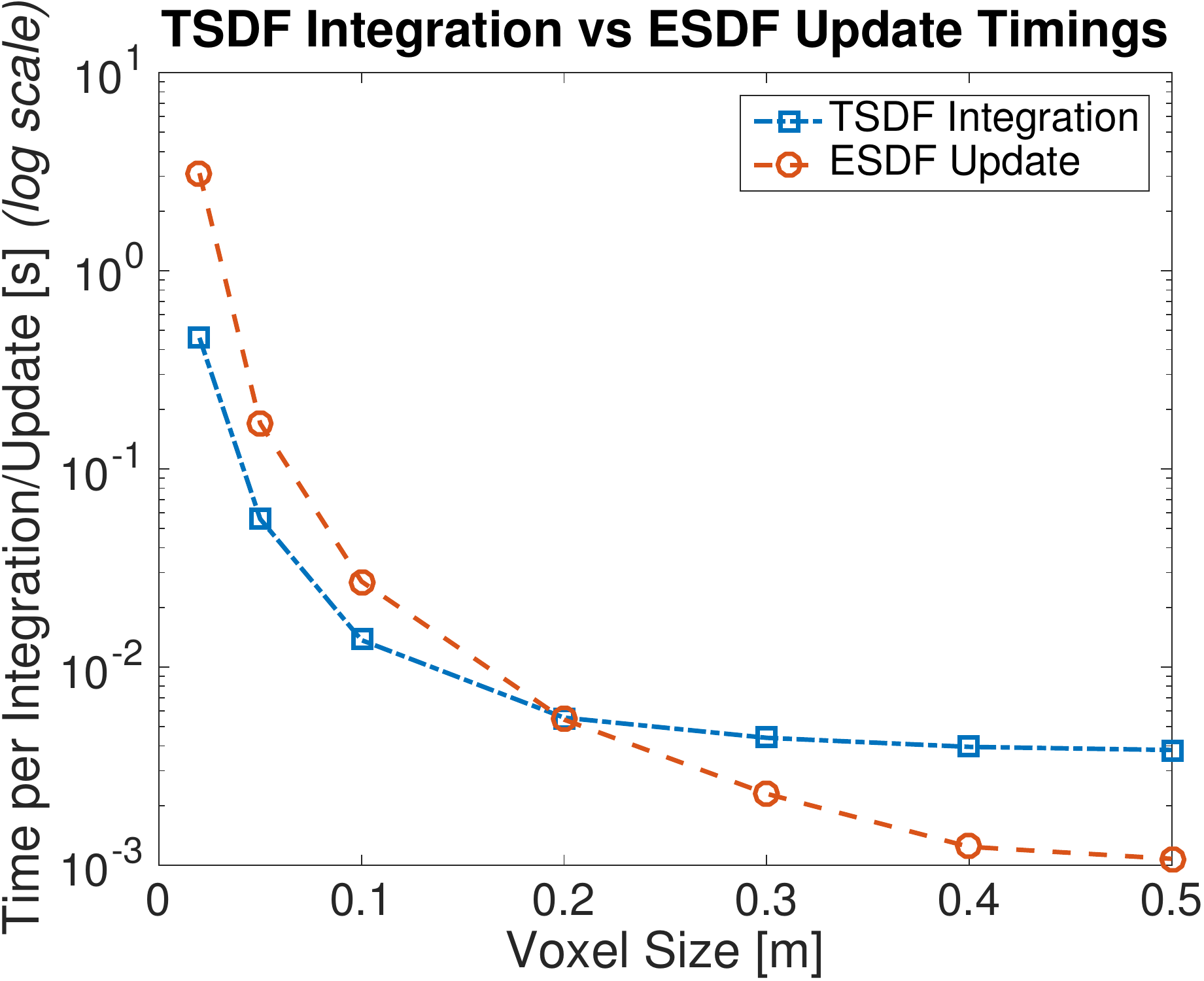}
  \caption{Timings results for integrating new data into the TSDF compared to propagating new TSDF updates to the ESDF on the EuRoC dataset. At small voxel sizes, TSDF integration is faster, but flattens out at large voxel sizes as the amount of sensor data does not decrease, while ESDF timings continue to decrease.}
  \label{fig:esdf_tsdf_timing}
\end{figure}

We also compare the integration time of the TSDF with update time of the ESDF layer in \reffig{fig:esdf_tsdf_timing}.
Though for small voxel sizes, the ESDF update is slower than integrating new TSDF scans, at large enough voxels (here, $v = 0.20$ m), the TSDF integration time flattens out while the ESDF update time keeps decreasing.
Since the number of points that need to be integrated into the TSDF does not vary with the voxel size, projecting the points into the voxel map dominates the timings for large voxels.

Based on these results, we recommend to use a single-voxel fixed band, quasi-Euclidean distance, and a priority queue for ESDF construction.

\section{MAV Planning Experiments}
\label{sec:mav_experiments}
To prove the usefulness of our generated ESDF for a real planning application, we set up an experiment with an MAV exploring an unknown space and replanning online as its ESDF gets updated.
A photo and screenshot of this experiment is shown in \reffig{fig:experiment}, and the complete trial can be seen in the video attachment.
The platform we used is an Asctec Firefly, equipped with a forward-facing stereo camera synced to an IMU, which we use for stereo matching as input to the mapping process, and also as input to the visual-inertial state estimator.
All state estimation, reconstruction, planning, and control runs entirely on-board on the Intel i7 2.1 GHz CPU without any reliance of any external equipment or infrastructure.

For this experiment, we use the continuous-time trajectory optimization replanning approach presented in \cite{oleynikova2016continuous-time}, which relies on having smooth collision costs and gradients from a Euclidean distance map.
We update the ESDF at 4 Hz, and replan after every map update, using 0.20 meter voxels.
Even with random restarts in the optimization procedure, the complete system including TSDF construction, ESDF updates, and replanning was able to run well under the 250 ms time budget.

We made one extension to the planning method to guarantee good performance: since the planner can not handle unknown space (as there are no collision gradients available), we allocate a $5$ meter sphere around the robot's current position and mark all unknown voxels in that sphere as occupied.
To compensate for fact that the MAV can not observe its current position (and would therefore mark it as unknown), we take a smaller $1$ meter sphere around the robot's start position and mark all unknown voxels in this smaller sphere as free.
Note that these changes do not affect any voxels that have actually been \textit{observed} -- only \textit{unknown} voxels are modified.

This experiment demonstrates that the proposed mapping approach can be used in combination with a planner and state estimator to navigate a small aerial robotic platform to a waypoint in a previously unknown environment while continually replanning to avoid obstacles. This is achieved while staying within the computational limits of the platform and operating in real time.  

\section{Conclusions}
This paper aims to find a suitable map representation for local planning on MAVs in unexplored environments.
Euclidean Signed Distance Fields (ESDFs) provide distance information to obstacles, which is essential for trajectory optimization planners.
In contrast, Truncated Signed Distance Fields (TSDFs) are fast to build, filter out noise in sensor data, and can be used to easily create meshes, which can then be interpreted by human operators.
We propose to incrementally build ESDFs directly out of TSDFs, rather than occupancy-based representations.
We extend existing methods to take advantage of distance information in the TSDF and allow dynamically-growing maps by using voxel hashing as the underlying data structure.

We focus on reducing the computational complexity of building these maps, while quantifying the errors introduced in our approximations to guarantee planning safety.
Our results suggest building a TSDF with 20 cm voxels, using grouped raycasting, and quadratic weights with linear drop-off behind a surface boundary. We also recommend using a one-voxel fixed band from the TSDF in order to build the ESDF, using quasi-Euclidean distances, and a distance-based priority queue for processing the open set.
Given possible sources of error in the maps, we recommend inflating the robot bounding box by $8.5\% + 0.3 v$, where $v$ is the voxel size.

We show that our method of building the TSDF is faster than building an Octomap, and that the accuracy of an ESDF built from a TSDF is higher than if built from an occupancy map.
Finally, we validate our complete system by building maps and using them to plan online with a trajectory optimization replanner, entirely on-board an MAV.

\bibliographystyle{ieeetr}

\bibliography{sources}

\end{document}